\documentclass[11pt]{article}

\usepackage[final]{acl}

\usepackage{times}
\usepackage{latexsym}

\usepackage[T1]{fontenc}

\usepackage[utf8]{inputenc}

\usepackage{microtype}

\usepackage{inconsolata}

\usepackage{graphicx}

\usepackage{booktabs}
\usepackage{amsmath}
\usepackage{multirow}
\usepackage{colortbl} 
\usepackage{xcolor}   

\definecolor{lightpink}{rgb}{0.98, 0.92, 0.92}

\title{ARCQuant: Boosting NVFP4 Quantization with Augmented Residual Channels for LLMs}

\author{
  Haoqian Meng\textsuperscript{1}, Yilun Luo\textsuperscript{1}, Yafei Zhao\textsuperscript{1}, Wenyuan Liu\textsuperscript{1}, Peng Zhang\textsuperscript{1}\thanks{\ \ Corresponding author: \texttt{pzhang@tju.edu.cn}}, Xindian Ma\textsuperscript{1} \\
  \textsuperscript{1}School of Computer Science and Technology, Tianjin University
}

\begin{document}
\maketitle
\begin{abstract}
The emergence of fine-grained numerical formats like NVFP4 presents new opportunities for efficient Large Language Model (LLM) inference. However, it is difficult to adapt existing Post-Training Quantization (PTQ) strategies to these formats: rotation-based methods compromise fine-grained block isolation; smoothing techniques struggle with significant 4-bit quantization errors; and mixed-precision approaches often conflict with hardware constraints on unified-precision computation. To address these challenges, we propose ARCQuant, a framework that boosts NVFP4 performance via Augmented Residual Channels. Distinct from methods that compromise block isolation or hardware uniformity, ARCQuant maintains a strictly unified NVFP4 format by augmenting the activation matrix with quantized residual channels. This design integrates the error compensation process directly into the matrix reduction dimension, enabling the use of standard, highly optimized GEMM kernels with minimal overhead. Theoretical analysis confirms that the worst-case error bound of our dual-stage NVFP4 quantization is comparable to that of standard 8-bit formats such as MXFP8. Extensive experiments on LLaMA and Qwen models demonstrate that ARCQuant achieves state-of-the-art accuracy, comparable to full-precision baselines in perplexity and downstream tasks. Furthermore, deployment on RTX 5090 and RTX PRO 6000 GPUs confirms practical benefits, achieving up to 3× speedup over FP16. Our code is available at \url{https://github.com/actypedef/ARCQuant}.  
\end{abstract}

\section{Introduction}

\label{sec:introduction}



\begin{figure*}[t]
    \centering
    \includegraphics[width=\textwidth]{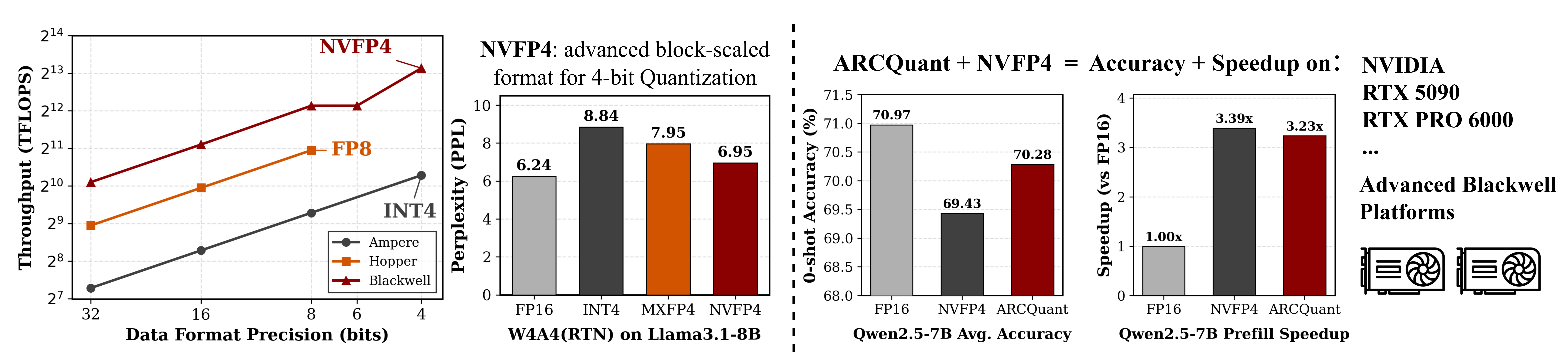}
  \caption{ARCQuant closes the NVFP4 accuracy gap while preserving high throughput on Blackwell platforms.}
  \label{fig:intro}
\end{figure*}

The deployment of Large Language Models (LLMs) faces severe constraints in memory bandwidth and computational latency.
To address these bottlenecks, Post-Training Quantization (PTQ) has established itself as a standard optimization strategy~\citep{xiao2024smoothquantaccurateefficientposttraining, frantar2023gptqaccurateposttrainingquantization}.
While 8-bit quantization and weight-only compression are widely adopted to reduce memory footprints~\citep{dettmers2022llmint88bitmatrixmultiplication, lin2023awq}, maximizing inference throughput requires quantizing both weights and activations to 4 bits (W4A4).
However, it remains a formidable challenge to maintain model fidelity under such aggressive compression.
Recently, hardware advancements such as the NVIDIA Blackwell architecture have introduced native support for fine-grained Microscaling formats, notably MXFP4 and NVFP4, as shown in Figure~\ref{fig:intro}~\citep{BlackwellArchitectureTechnical, rouhani2023microscalingdataformatsdeep, nvidia2024nvfp4}.

Microscaling formats like NVFP4 represent a significant advancement in efficient deep learning, demonstrating effectiveness in low-precision training and high-throughput inference~\citep{chmiel2025fp4, nvidia2024nvfp4}.
Their primary advantage over coarse-grained quantization lies in fine-grained block-wise isolation, which prevents high-magnitude outliers from inflating scaling factors across entire tensors.
However, it is insufficient to simply adopt this format for aggressive W4A4 constraints.
Existing PTQ strategies often fail to exploit this structure, necessitating specialized algorithms that optimize quantization error while respecting NVFP4's unique block-wise properties.

Most W4A4 strategies designed for integer formats are ill-suited for the specific constraints of fine-grained NVFP4.
Global transformations often prove counterproductive for block-wise scaling by redistributing outlier magnitudes across dimensions.
This propagation inflates local dynamic ranges, thereby compromising the isolation capability inherent to the format.
Furthermore, mixed-precision strategies such as Atom encounter significant hardware barriers.
The NVFP4 standard requires a block size of 16, creating a structural incompatibility with the coarser granularities of higher-precision formats.
This discrepancy precludes the use of optimized Tensor Core instructions, which require unified data paths for high-throughput execution.

To overcome these algorithmic and hardware barriers, we propose ARCQuant (Augmented Residual Channels Quantization). 
Unlike mixed-precision approaches that violate hardware uniformity, ARCQuant maintains a strictly unified NVFP4 data path. 
Our method identifies dominant outlier channels and augments the input tensor with their quantized residuals. 
This facilitates a dual-stage quantization mechanism: the primary stage captures the high-magnitude structure, while the augmented stage recovers the fine-grained residual information. 
Crucially, this design maps the entire compensation process into the extended reduction dimension of a single matrix multiplication. 
Consequently, ARCQuant leverages standard and highly optimized GEMM kernels to achieve W4A8-level accuracy, while adhering to W4A4 hardware constraints.

Our contributions are summarized as follows:
\begin{itemize}
    \item We propose ARCQuant, the first PTQ framework optimized for NVFP4 that utilizes augmented residual channels to bypass mixed-precision hardware constraints. Extensive experiments on LLaMA and Qwen families demonstrate that ARCQuant consistently outperforms existing NVFP4 adaptation strategies, delivering W4A8-level performance across various model families.
    \item We provide a rigorous analysis demonstrating that the worst-case error bound of our dual-stage mechanism is comparable to that of standard single-stage MXFP8. This theoretically validates our capability to bridge the precision gap between W4A4 and W4A8.
    \item We implement high-performance kernels that map the complex residual accumulation process into standard, unified-precision GEMM calls, ensuring portability and efficiency. On RTX 5090 and PRO 6000, we achieve significant improvements in prefill speed and memory efficiency compared to FP16.
\end{itemize}

\section{Related Work}

\label{sec:related_work}


\begin{figure*}[t]
    \centering
    \includegraphics[width=\textwidth]{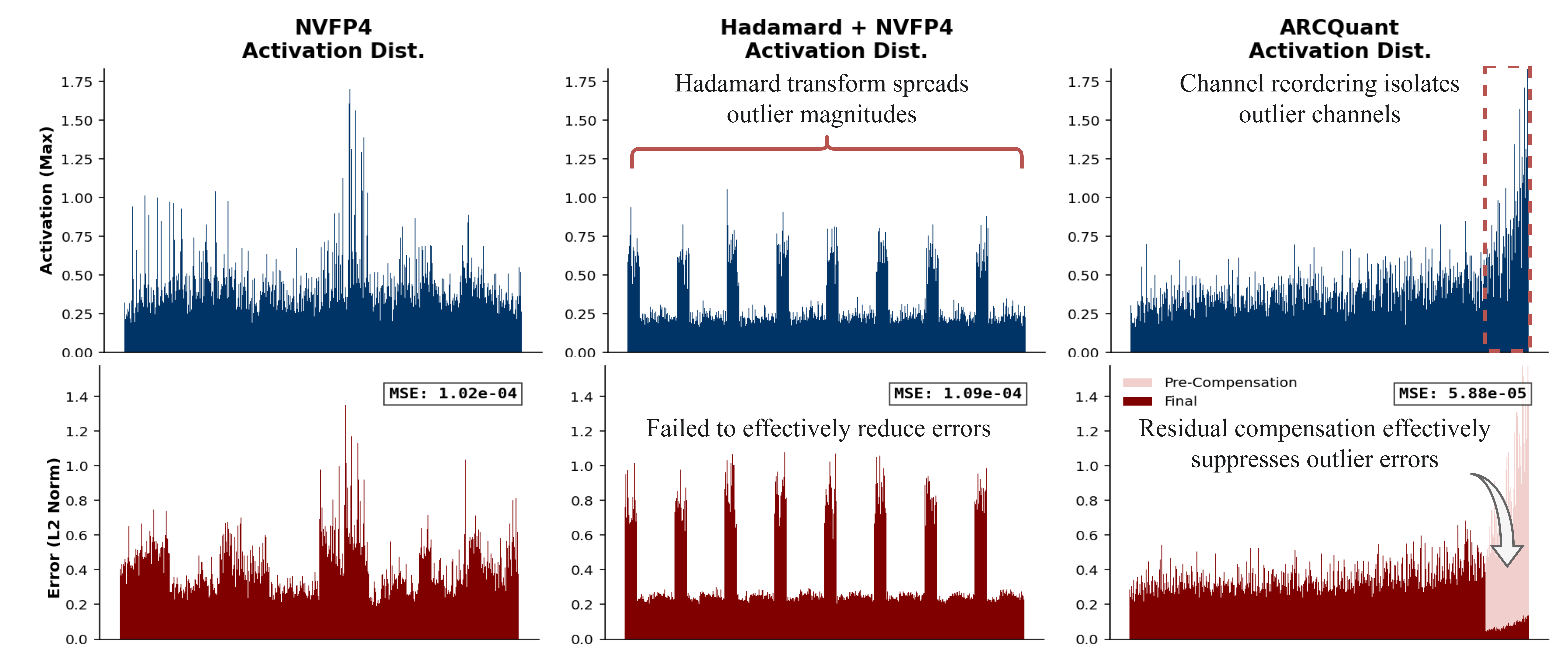} 
    \caption{
        Activation magnitudes (blue) and quantization errors (red) on Llama 3.1-8B \texttt{o\_proj}. ARCQuant isolates outlier to effectively suppress errors via residual compensation, whereas Hadamard spreads outlier magnitudes.
    }
    \label{fig:visualize}
\end{figure*}

Post-Training Quantization (PTQ) is pivotal for efficient LLMs. Early work identified activation outliers as the primary low-bit bottleneck~\citep{dettmers2022llmint88bitmatrixmultiplication}. While initial methods focused on weight-only quantization~\citep{lin2023awq}, higher compression demands have shifted focus to activation quantization and fine-grained formats like NVFP4. We categorize existing strategies into mixed-precision, transformation-based, and compensation-based methods.

\paragraph{Mixed-precision Methods.} These approaches mitigate quantization error by retaining sensitive information in higher precision.
Early methods like Atom~\citep{atom} and QUIK~\citep{QUIK} adopt a selection-based strategy, keeping the majority of weights and activations in INT4 while preserving a subset of outlier channels in INT8 or FP16.
Beyond simple channel selection, decomposition-based approaches such as ResQ~\citep{saxena2025resqmixedprecisionquantizationlarge} and SVDQuant~\citep{li2024svdquant} extract outliers into a separate high-precision low-rank branch, leaving the residual bulk in 4-bit.
Recent works have extended mixed-precision to emerging hardware formats.
MicroMix~\citep{liu2025micromix} explores mixing MXFP4 and MXFP8 on the NVIDIA Blackwell architecture.
Similarly, FGMP~\citep{hooper2025fgmpfinegrainedmixedprecisionweight} proposes a strategy combining NVFP4 with FP8; while currently lacking native hardware support, it outlines a viable direction for future architectural designs.

\paragraph{Transformation-based Methods.} These strategies reshape distributions to suppress outliers. SmoothQuant~\citep{xiao2024smoothquantaccurateefficientposttraining} migrates quantization difficulty to weights, proving effective for INT8. For 4-bit scenarios, methods like QUIP~\citep{tseng2024quip}, QuaRot~\citep{ashkboos2024quarot}, and FlatQuant~\citep{sun2024flatquant} employ randomized rotations or learnable affine transformations to flatten activation distributions. However, the effectiveness of such global transformations on fine-grained formats is contested. Recent studies like BRQ~\citep{shao2025block} indicate that these operations may disrupt local block statistics—a critical limitation for NVFP4 that we further analyze in Section~\ref{sec:methodology}.

\paragraph{Compensation-based Methods.}
Strategies such as GPTQ~\citep{frantar2023gptqaccurateposttrainingquantization} and APTQ~\citep{guan2024aptq} minimize reconstruction error by utilizing Hessian-based optimization to adjust quantized weights.
While highly effective for static parameter compression, these methods do not address the dynamic, runtime compensation of activation quantization errors that ARCQuant targets.

\begin{figure}[t]
  \centering
  \includegraphics[width=\columnwidth]{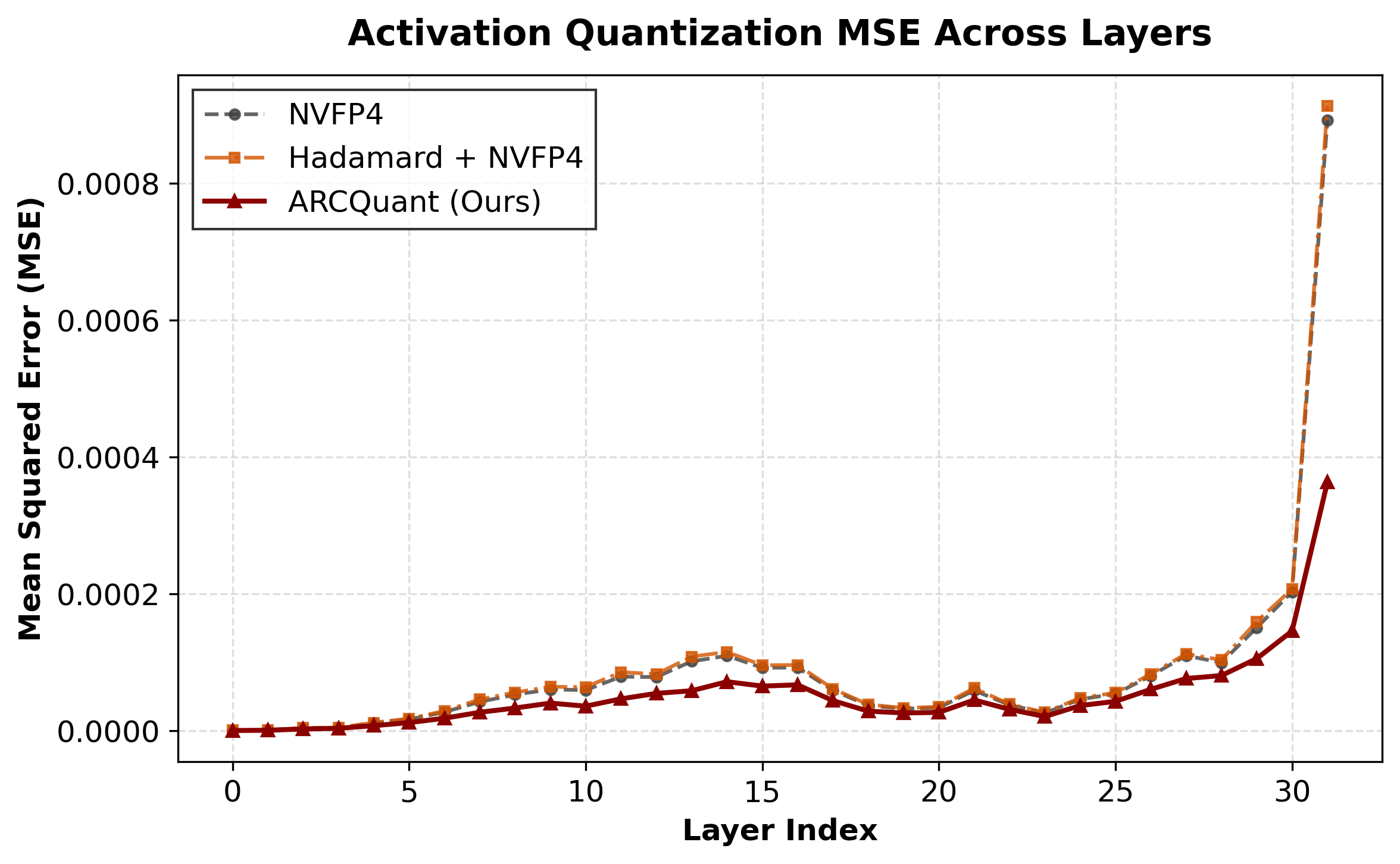}
  \caption{\texttt{o\_proj} in Llama 3.1-8B, ARCQuant consistently suppressed MSE on NVFP4 across all layers. }
  \label{fig:mse}
\end{figure}

\section{Methodology}

\label{sec:methodology}

\subsection{Preliminary and Motivation}

\paragraph{Problem Definition.} We consider the linear layer computation $Y = XW^\top$. Following the notation in QServe~\citep{lin2025qservew4a8kv4quantizationcodesign}, we denote the configuration with $x$-bit weights and $y$-bit activations as W$x$A$y$. ARCQuant specifically targets the W4A4 setting. Let $g$ denote the quantization group size and $q_{\text{max}}$ be the maximum representable value. For each group, the low-bit representation $Q_X$ and the scaling factor $s_X$ are calculated as:
\begin{equation}
\begin{aligned}
    Q_X &= \text{round}\left(\frac{X}{s_X}\right), \\
    \quad s_X &= \frac{\max(|X|)}{q_{\text{max}}}.
\end{aligned}
\end{equation}
The dequantized approximation is given by $Q(X) = s_X \cdot Q_X$. Our objective is to minimize the reconstruction error $\|Y - Q(X)Q(W)^\top\|_F$.

\begin{figure*}[t]
    \centering
    \includegraphics[width=\textwidth]{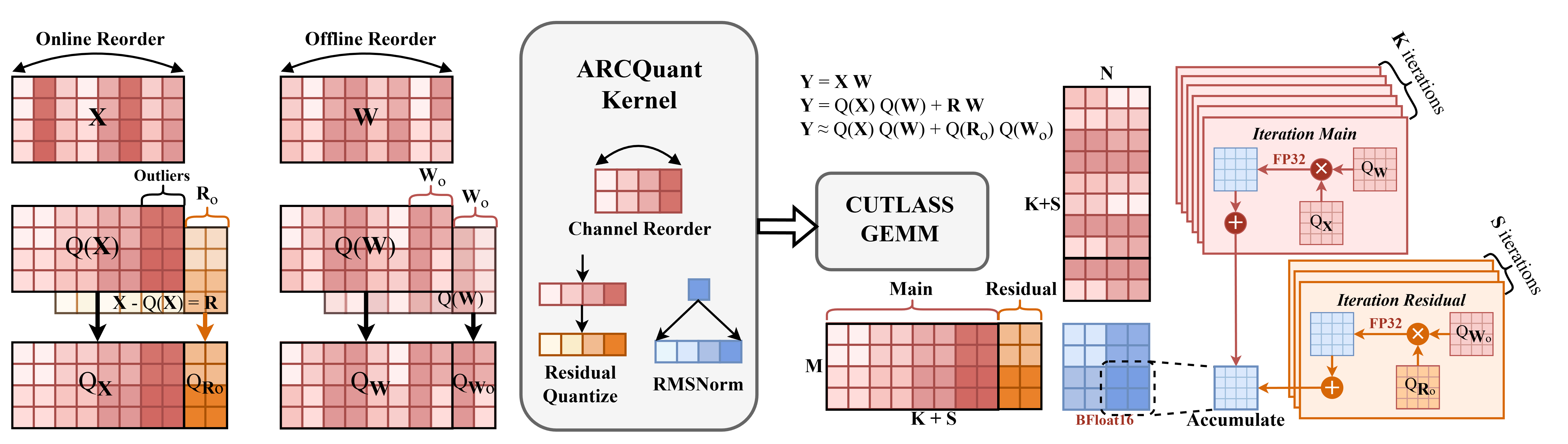} 
    \caption{ARCQuant fuses main and residual computations by mapping them into the extended dimension. }
    \label{fig:kernels}
\end{figure*}

\paragraph{Block-scaled Quantization.} NVIDIA Blackwell architecture introduces hardware acceleration for block-scaled formats. Unlike per-tensor methods, quantization is performed on small groups. Specifically, MXFP4 ($g=32$) comprises 32 E2M1 elements sharing one E8M0 scaling factor, whereas NVFP4 ($g=16$) uses 16 E2M1 elements with a shared E4M3 scale (details in Appendix~\ref{appendix:block_formats}). Due to the limited dynamic range of E4M3, NVFP4 requires a secondary per-tensor scaling factor. This fine-grained approach isolates outliers within specific blocks, preventing maximum values in outlier channels from inflating the scaling factors of adjacent low-magnitude elements. ARCQuant is specifically designed to leverage this isolation property to mitigate quantization errors arising from activation outlier channels.

\paragraph{Limitations of Global Rotations.} Rotation strategies typically aim to suppress outliers by flattening the magnitude distribution across channels. While this reduces the global row-wise maximum, we observe it is detrimental for fine-grained NVFP4. As shown in Figures~\ref{fig:visualize} and~\ref{fig:mse}, the Hadamard transform involves a linear combination that propagates the high magnitude of outlier channels into all dimensions. While the global peak decreases, this operation significantly increases the local dynamic range of previously low-magnitude blocks. This effect negates the isolation benefit of fine-grained scaling. In contrast, ARCQuant isolates outlier channels via reordering, performing targeted compensation while preserving the original numerical values of the remaining elements.

\paragraph{Hardware Constraints on Mixed-Precision.} Conventional mixed-precision strategies like Atom protect sensitive channels using higher precision formats such as INT8 or FP16. However, this approach is inefficient for Blackwell NVFP4 acceleration. While NVFP4 operates with a group size of $g=16$, standard higher-precision formats like MXFP6 and MXFP8 are designed for $g=32$. Crucially, current Tensor Core instructions do not support efficient Matrix Multiply-Accumulate (MMA) operations on operands with heterogeneous group sizes. Mixing granularities precludes the use of these hardware-accelerated pipelines, necessitating complex kernel logic for memory alignment and significantly degrading inference throughput.

\paragraph{Motivation.}
These limitations present a critical dilemma: Global rotations propagate outlier magnitudes, while mixed-precision formats violate hardware uniformity. This raises the question: How can we mitigate outlier errors in fine-grained NVFP4 without these drawbacks? To this end, we propose \textit{ARCQuant}. Instead of transforming input values or mixing data formats, ARCQuant augments the linear layer with residual channels of identical W4A4 precision. This design allows targeted error compensation via the standard NVFP4 GEMM kernels, effectively achieving high-fidelity reconstruction while strictly adhering to hardware constraints.

\subsection{Augmented Residual Channels Strategy}

\begin{figure}[t]
  \centering
  \includegraphics[width=\columnwidth]{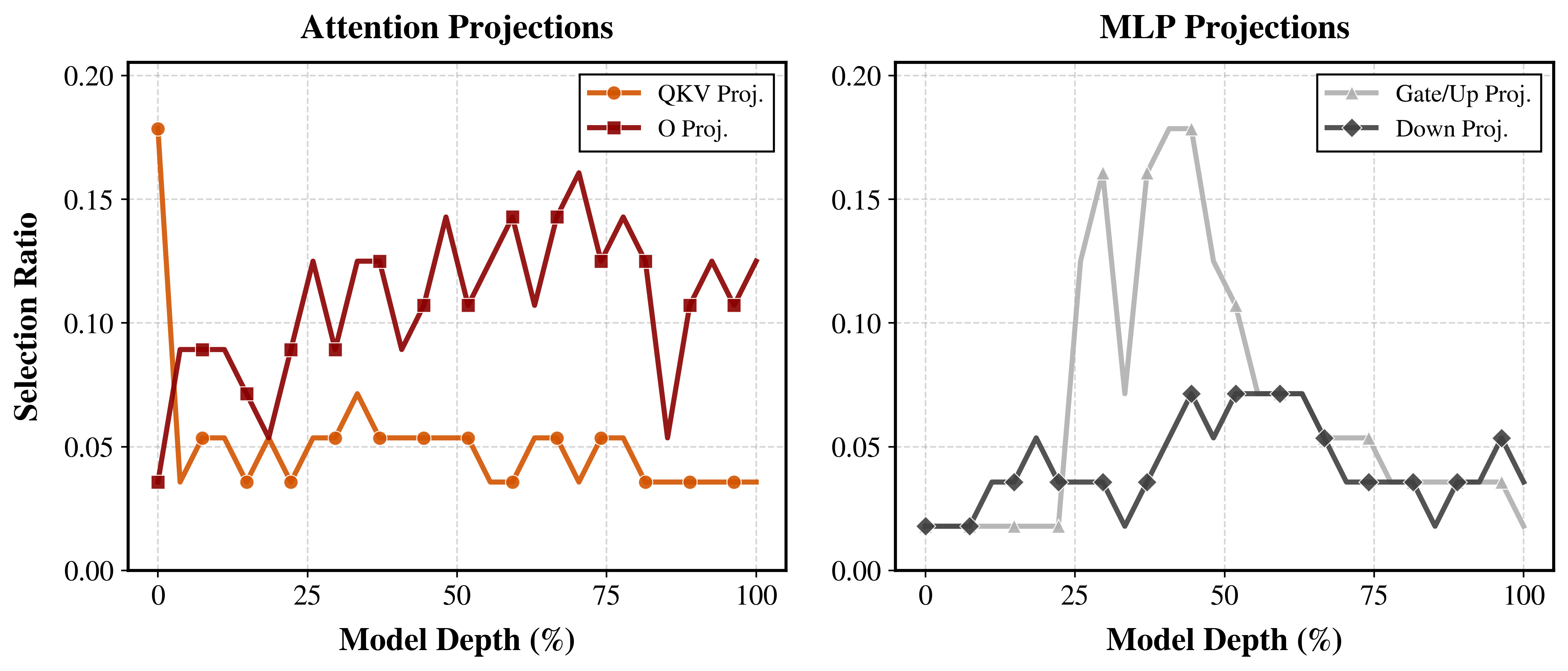}
  \caption{Outlier channels $S$ across Qwen2.5-7B layers.}
  \label{fig:layers}
\end{figure}

\paragraph{Adaptive Outlier Identification.} We aim to maximize reconstruction accuracy with minimal cost by identifying channels that strictly require compensation. Utilizing calibration data, we pre-determine both the channel reordering indices and the number of outlier channels $S$. We first reorder channels based on their absolute maximums, adopting the sorting strategy from Atom~\citep{atom}. We then determine the layer-wise maximum $M$, which defines the dynamic range of a per-tensor FP8 (E5M2) reference. We set the selection threshold $\tau = 2^{-3} M$. This reflects the 3-bit difference in exponent width between the reference E5M2 (5 bits) and the target E2M1 (2 bits). Values below this threshold ($|x| \le \tau$) fall into the lower range of the FP8 format. In this range, the precision of NVFP4 is comparable to the baseline, making additional compensation unnecessary. Therefore, we focus solely on the top-$S$ reordered channels exceeding this boundary, where the precision gap is significant due to the limited dynamic range. Detailed statistics on the layer-wise assignment of $S$ on different models are shown in Figure~\ref{fig:layers}.

\paragraph{Online Activation Quantization.} We process $X$ via: (1) \textit{Reordering and Primary Quantization.} $X$ is reordered, followed by block-wise quantization $Q_X = \text{round}(X / s_X)$. (2) \textit{Residual Compensation.} We isolate outliers $X_{\text{o}}$, compute residuals $R_{\text{o}} = X_{\text{o}} - s_{X_{\text{o}}} \cdot Q_{X_{\text{o}}}$, and quantize them into $Q_{R_{\text{o}}}$. (3) \textit{Augmentation.} We concatenate along dimension $K$ as $Q_{X_{\text{aug}}} = [Q_X \mid Q_{R_{\text{o}}}]$ with combined scales $s_{X_{\text{aug}}} = [s_X \mid s_{R_{\text{o}}}]$.

\paragraph{Offline Weight Quantization.} Weights are aligned offline via: (1) \textit{Reordering.} $W$ is reordered to match $X$ and quantized into $Q_W$. (2) \textit{Augmentation.} Instead of computing residuals, we duplicate the quantized outlier weights $Q_{W_{\text{o}}}$. We construct $Q_{W_{\text{aug}}} = [Q_W \mid Q_{W_{\text{o}}}]$, ensuring the GEMM effectively computes the correction term $R_{\text{o}} Q(W_{\text{o}})^\top$.

\paragraph{Unified GEMM Execution.} The original matrix multiplication $(N, K_{\text{in}}, M)$ is transformed into an augmented operation $(N, K_{\text{in}} + S, M)$.
The mathematical equivalence is derived as follows:
\begin{equation}
\begin{aligned}
Y &\approx Q(X)Q(W)^\top + Q(R_{\text{o}})Q(W_{\text{o}})^\top \\
  &= s_{X_{\text{aug}}} \cdot Q_{X_{\text{aug}}} (s_{W_{\text{aug}}} \cdot Q_{W_{\text{aug}}})^\top. 
\end{aligned}
\end{equation}
By constructing $Q_{X_{\text{aug}}}$ and $Q_{W_{\text{aug}}}$ in the same precision, the compensation is seamlessly integrated into a single GEMM call.

\subsection{ARCQuant Kernel Design}

\begin{figure}[t]
    \centering
    \includegraphics[width=\columnwidth]{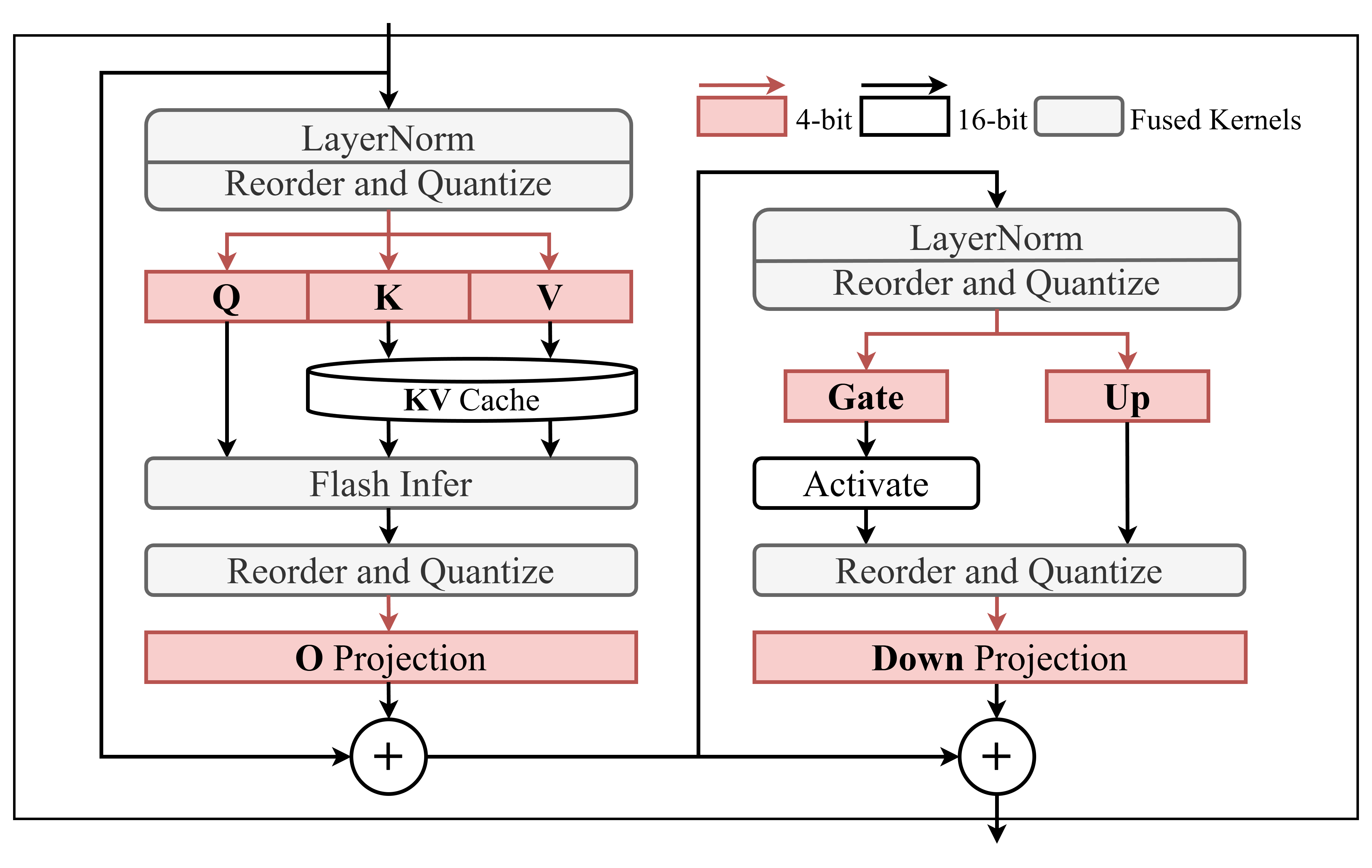} 
    \caption{ARCQuant on a Transformers block in LLM. }
    \label{fig:transformers}
\end{figure}

While compensation-based weight quantization incurs one-time offline costs, ARCQuant requires dynamic, online residual calculation. Achieving this efficiently poses significant challenges for CUDA kernel design. To address this, we implement a Fused Quantization Kernel that integrates Channel Reordering, RMSNorm, Primary Quantization, and Residual Quantization into a single operation. We utilize coalesced memory access patterns during reordering to minimize latency, and output data in the exact Interleaved Channel Layout (detailed in Appendix~\ref{appendix:kernel_design}) for the subsequent GEMM.

The output of this fused kernel is an augmented tensor strictly adhering to the NVFP4 format. As matrix multiplication dominates the computational load, this allows us to leverage standard, optimized CUTLASS GEMM kernels without modification, as shown in Figures~\ref{fig:kernels} and~\ref{fig:transformers}. This design ensures hardware-native efficiency and simplifies deployment. 
Mechanically, the GEMM operates on the extended reduction dimension ($K_{\text{in}} + S$). Due to the linearity of the accumulation, it seamlessly sums the primary computation (first $K_{\text{in}}$ channels) and the residual corrections (final $S$ channels) directly into the high-precision output accumulator.







\subsection{Error Bound Analysis}

Our analysis confirms that, for compensated outlier channels, the dual-stage mechanism matches the worst-case error bounds of standard MXFP8.

\paragraph{Preliminaries.} Let $M$ be the dynamic range. The worst-case quantization error is bounded by $|e| \le s \epsilon = \alpha M \epsilon$, where $\alpha = s / M \ge 1$ is the scale alignment overhead and $\epsilon$ is the precision limit. Specifically, we consider NVFP4 (E2M1, $\epsilon_4 = 2^{-2}$, scale E4M3) and MXFP8 (E4M3, $\epsilon_8 = 2^{-4}$, scale E8M0). Since $\epsilon_4^2 = \epsilon_8$, a dual-stage NVFP4 mechanism theoretically bridges the precision gap, matching the resolution of standard MXFP8.

\paragraph{Derivation of Error Bounds.} For MXFP8, the E8M0 scale (powers of 2) implies an alignment factor $\alpha_{mx} \in [1, 2)$. The worst-case bound is:
\begin{equation}
    B_{\text{mx}} = \alpha_{\text{mx}} M \epsilon_8 < 2 M \epsilon_8. 
\end{equation}

For ARCQuant, we use a two-stage process on outliers:
(1) quantize $\mathbf{x}$ with scale $s_1 = \alpha_1 M$. The residual is bounded by $\|\mathbf{r}\|_\infty \le \alpha_1 M \epsilon_4$. (2) then quantize the residual with scale $s_2 = \alpha_2 \|\mathbf{r}\|_\infty$. Substituting $\epsilon_4^2 = \epsilon_8$, the bound becomes: 
\begin{equation}
\begin{split}
    |e_{\text{arc}}|
&\le s_2 \epsilon_4 \le (\alpha_2 \alpha_1 M \epsilon_4) \epsilon_4 \\
              &= (\alpha_1 \alpha_2) M \epsilon_8 = B_{\text{arc}}. 
\end{split}
\end{equation}

\begin{table*}[t]
\centering
\caption{Zero-shot, few-shot accuracy and perplexity. \textbf{Bold} indicates the best performance among the compared W4A4 methods.}
\label{tab:main_results}
\begin{tabular}{l|cccccc|c|c} 
\toprule
 & \multicolumn{6}{c|}{\textbf{Zero-shot Accuracy} ($\uparrow$)} & \textbf{PPL} ($\downarrow$) & \textbf{MMLU} ($\uparrow$) \\
\textbf{Method} & Arc-C & Hella & Lamba & PIQA & Wino & \textbf{Average} & WikiText2 & 5-shot \\
\midrule
\multicolumn{9}{c}{\textbf{Llama 3.1-8B}} \\
\midrule
FP16 & 53.41 & 78.91 & 75.28 & 81.34 & 73.88 & 72.56 & 6.24 & 65.15 \\
W4A8 + RTN    & 50.51 & 77.62 & 73.16 & 79.82 & 71.82 & 70.59 & 7.07 & 61.08 \\
\cmidrule(lr){1-9} 
FlatQuant & 51.54 & 76.56 & 73.32 & 79.16 & 71.98 & 70.51 & 6.95 & 61.33 \\
MicroMix      & 48.38 & 76.04 & 72.15 & 79.27 & 70.56 & 69.28 & 7.35 & 60.17 \\
Atom      & 47.53 & 74.22 & 69.45 & 78.02 & 69.46 & 67.74 & 7.52 & 59.27 \\
\rowcolor{lightpink} \textbf{ARCQuant} & 52.22 & 77.22 & 72.79 & 79.16 & 73.09 & \textbf{70.90} & \textbf{6.87} & \textbf{62.61} \\

\midrule \midrule 

\multicolumn{9}{c}{\textbf{Qwen2.5-7B}} \\
\midrule
FP16 & 51.28 & 78.98 & 71.63 & 79.71 & 73.24 & 70.97 & 6.85 & 74.16 \\
W4A8 + RTN    & 51.19 & 77.30 & 67.22 & 79.71 & 70.32 & 69.15 & 7.44 & 71.88 \\
\cmidrule(lr){1-9}
FlatQuant & 50.00 & 76.59 & 68.81 & 79.43 & 71.03 & 69.17 & 7.88 & 71.95 \\
MicroMix      & 51.28 & 76.45 & 69.30 & 79.49 & 68.82 & 69.07 & 7.69 & 70.99 \\
Atom      & 48.38 & 74.63 & 68.62 & 77.64 & 68.59 & 67.57 & 8.96 & 68.17 \\
\rowcolor{lightpink} \textbf{ARCQuant} & 51.79 & 77.88 & 70.68 & 79.05 & 71.98 & \textbf{70.28} & \textbf{7.28} & \textbf{72.84} \\

\midrule \midrule

\multicolumn{9}{c}{\textbf{Qwen2.5-32B}} \\
\midrule
FP16 & 55.89 & 84.17 & 76.17 & 82.43 & 75.45 & 74.82 & 5.02 & 83.26 \\
W4A8 + RTN    & 57.94 & 83.10 & 74.79 & 82.43 & 76.72 & 75.00 & 5.41 & 82.57 \\
\cmidrule(lr){1-9}
FlatQuant & 56.23 & 82.79 & 75.41 & 81.50 & 74.19 & 74.02 & 5.74 & 81.52 \\
MicroMix      & 56.57 & 82.92 & 73.04 & 82.10 & 73.56 & 73.64 & 5.69 & 81.75 \\
Atom      & 54.78 & 82.38 & 75.92 & 81.45 & 73.48 & 73.60 & 5.83 & 79.54 \\
\rowcolor{lightpink} \textbf{ARCQuant} & 56.57 & 83.36 & 76.05 & 82.64 & 75.37 & \textbf{74.80} & \textbf{5.38} & \textbf{82.61} \\
\bottomrule
\end{tabular}
\end{table*}

\paragraph{Comparison.} We assess theoretical parity by comparing alignment factors: MXFP8 uses exponent-only E8M0 scales ($\sup \alpha_{mx} = 2$), whereas NVFP4 employs mantissa-coded E4M3 scales ($2^{-3}$ step size) yielding $\sup \alpha_1 \alpha_2 = 1.125^2 \approx 1.266$. Since $1.266 < 2$, ARCQuant achieves a worst-case error bound competitive with MXFP8, confirming that the dual-stage mechanism effectively restores high-fidelity representation for sensitive channels within strict W4A4 constraints.

\begin{table*}[t]
\centering
\caption{Zero-shot, few-shot accuracy and perplexity. \textbf{Bold} indicates the best performance among the NVFP4 methods.}
\label{tab:nvfp4}
\begin{tabular}{l|cccccc|c|c}
\toprule
 & \multicolumn{6}{c|}{\textbf{Zero-shot Accuracy} ($\uparrow$)} & \textbf{PPL} ($\downarrow$) & \textbf{MMLU} ($\uparrow$) \\
\textbf{Method} & Arc-C & Hella & Lamba & PIQA & Wino & \textbf{Average} & WikiText2 & 5-shot \\
\midrule
\multicolumn{9}{c}{\textbf{Llama 3.1-8B}} \\
\midrule
NVFP4 + RTN         & 50.26 & 77.56 & 73.84 & 78.67 & 71.90 & 70.45 & 6.95 & 61.64 \\
NVFP4 + Smooth & 50.34 & 77.51 & 73.32 & 79.22 & 71.59 & 70.40 & 6.92 & 61.76 \\
NVFP4 + QuaRot      & 51.37 & 77.40 & 72.64 & 79.00 & 70.48 & 70.18 & 6.99 & 61.73 \\
\rowcolor{lightpink} \textbf{ARCQuant} & 52.22 & 77.22 & 72.79 & 79.16 & 73.09 & \textbf{70.90} & \textbf{6.87} & \textbf{62.61} \\

\midrule \midrule 

\multicolumn{9}{c}{\textbf{Qwen2.5-7B}} \\
\midrule
NVFP4 + RTN         & 51.19 & 77.55 & 70.37 & 78.73 & 69.30 & 69.43 & 7.29 & 72.06 \\
NVFP4 + Smooth & 51.62 & 77.56 & 70.70 & 79.05 & 69.85 & 69.76 & 7.29 & 72.33 \\
NVFP4 + QuaRot      & 50.00 & 77.52 & 70.56 & 79.22 & 70.01 & 69.46 & 7.30 & 72.33 \\
\rowcolor{lightpink} \textbf{ARCQuant} & 51.79 & 77.88 & 70.68 & 79.05 & 71.98 & \textbf{70.28} & \textbf{7.28} & \textbf{72.84} \\
\bottomrule
\end{tabular}
\end{table*}

\section{Experiments}

\label{sec:experiments}

\subsection{Experimental Setup}


\paragraph{Models and Datasets.}
We evaluate ARCQuant on a diverse set of models, including Llama 3.1-8B~\citep{grattafiori2024llama3herdmodels} and the Qwen2.5 family (7B, 32B, Coder-7B-Instruct, Math-7B-Instruct)~\citep{qwen2025qwen25technicalreport}. The metrics include: (1) \textit{PPL} on WikiText2~\citep{wikitext}; (2) \textit{Reasoning}: 5-shot MMLU~\citep{mmlu} and avg. 0-shot on ARC-C, HellaSwag, PIQA, Winogrande, Lambada~\citep{allenai:arc, zellers2019hellaswag, piqa, sakaguchi2019winograndeadversarialwinogradschema, lambada}, using lm-eval~\citep{lm-eval}; and (3) \textit{Domain Tasks}: HumanEval, MBPP, GSM8K, CMATH~\citep{humaneval, mbpp, cobbe2021gsm8k, wei2023cmath}. (See Appendix~\ref{appendix:details} for details).

\paragraph{Baselines.}

To ensure a comprehensive and fair evaluation, we benchmark ARCQuant against the FP16 baseline and a diverse set of state-of-the-art quantization methods using their official configurations.
Most baselines in our comparison operate in W4A4. In particular, we establish performance lower bounds using Round-To-Nearest (RTN) applied across NVFP4, MXFP4, and INT4 formats. We also include a higher-precision W4A8 reference implemented with MXFP4 weights and MXFP8 activations.
Beyond these fixed-precision baselines, we further evaluate MicroMix~\citep{liu2025micromix} as a mixed-precision microscaling method, which adopts MXFP4 weights and channel-wise MXFP4/MXFP8 activations.
For advanced PTQ frameworks, we evaluate FlatQuant~\citep{sun2024flatquant} and Atom~\citep{atom} using their original configurations, as their complex learning strategies and customized CUDA kernels are structurally incompatible with NVFP4. Additionally, we adapt SmoothQuant~\citep{xiao2024smoothquantaccurateefficientposttraining} and QuaRot~\citep{ashkboos2024quarot} to the NVFP4 format to assess their viability in fine-grained block-wise quantization.
Detailed implementation settings for all baselines are provided in Appendix~\ref{appendix:reproducibility}.

\paragraph{Hardware.} We conduct efficiency evaluations on NVIDIA RTX 5090 and RTX PRO 6000 GPUs, measuring both kernel-level latency and end-to-end inference throughput. Detailed statistics regarding post-quantization memory footprint and quantization latency are provided in Appendix~\ref{appendix:details}.

\subsection{Main Results}

\begin{table}[t]
\centering
\caption{Code generation performance on Qwen2.5-Coder-7B-Instruct. We report pass@1 accuracy on HumanEval, HumanEval+ (HE+), Mbpp, and Mbpp+.}
\label{tab:code_results}
\setlength{\tabcolsep}{4.5pt}
\begin{tabular}{l|cccc}
\toprule
 & \multicolumn{4}{c}{\textbf{Qwen2.5-Coder-7B-Instruct}} \\
\cmidrule(lr){2-5}
\textbf{Method} & HE & HE+ & Mbpp & Mbpp+ \\
\midrule
FP16 & 84.1 & 79.9 & 80.4 & 67.2 \\
\midrule
Atom & 80.5 & 76.2 & 74.5 & 63.2 \\
\rowcolor{lightpink} \textbf{ARCQuant} & \textbf{86.0} & \textbf{79.3} & \textbf{79.9} & \textbf{68.3} \\
\bottomrule
\end{tabular}
\end{table}


\paragraph{Accuracy Comparison.} Table~\ref{tab:main_results} reports the performance across Llama 3.1 and Qwen2.5 families. ARCQuant consistently achieves the best results among W4A4 methods, significantly outperforming Atom and FlatQuant. Notably, on Llama 3.1-8B and Qwen2.5-7B, ARCQuant surpasses the W4A8 RTN baseline in both perplexity and MMLU scores, effectively closing the gap to full precision. For instance, on Qwen2.5-7B, our method reduces perplexity by 1.68 points compared to Atom. On the larger Qwen2.5-32B, ARCQuant demonstrates near-lossless compression, matching the FP16 baseline within a negligible margin.

\paragraph{Quantization Strategies on NVFP4.} We evaluate four strategies: RTN, SmoothQuant, QuaRot, and ARCQuant. Table~\ref{tab:nvfp4} shows that conventional PTQ methods yield diminishing returns on NVFP4. Notably, QuaRot (Hadamard transform) results in performance regression compared to RTN on Llama 3.1-8B, which confirms our motivation (Figure~\ref{fig:visualize}) that rotation disrupts the isolation benefits of fine-grained blocking. Similarly, SmoothQuant offers marginal gains due to the limited capacity of 4-bit weights. In contrast, ARCQuant consistently achieves the best results in all metrics, effectively suppressing outlier errors in the activation matrices.


\subsection{Efficiency and Speed}

\begin{figure*}[t]
    \centering
    \includegraphics[width=\textwidth]{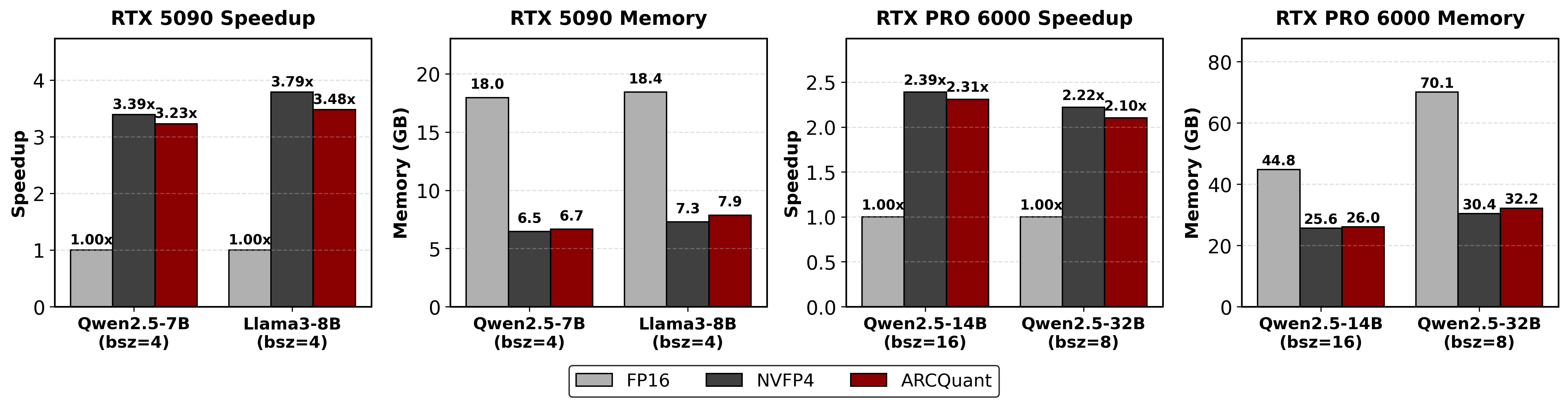} 
    \caption{Prefill efficiency for various models with 2048 sequence length on RTX 5090 and PRO 6000.}
    \label{fig:e2e}
\end{figure*}

\begin{figure}[t]
  \centering
  \includegraphics[width=\columnwidth]{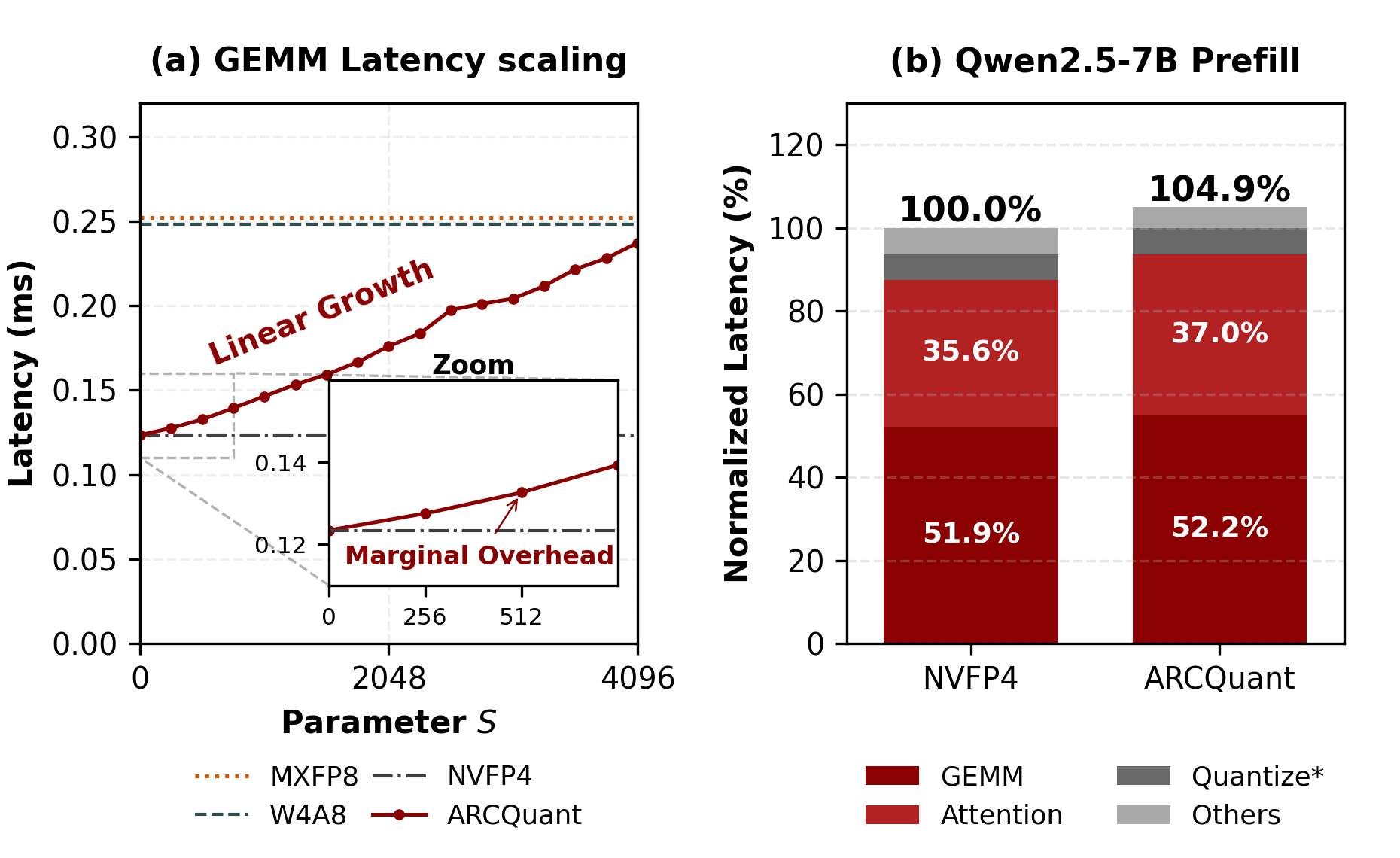}
  \caption{Efficiency Benchmarks. 
  \textbf{(a)} Kernel latency vs. channel count $S$. The inset ($S \le 512$) confirms ARCQuant is significantly faster than W4A8. 
  \textbf{(b)} Qwen2.5-7B prefill breakdown showing 4.9\% total overhead. 
*includes Reorder, RMSNorm, and Residual Quantize.}
  \label{fig:gemm_bench}
\end{figure}

\paragraph{Efficiency Analysis.} We analyze the computational overhead of ARCQuant through both kernel-level benchmarks and end-to-end latency profiling. Figure~\ref{fig:gemm_bench}\textbf{(a)} shows that the GEMM latency exhibits a strictly linear correlation with the number of augmented channels $S$. Crucially, the inset highlights that within the typical operating range ($S \le 512$), ARCQuant incurs only marginal overhead over NVFP4 baseline, while maintaining a significant speed advantage over W4A8 and MXFP8 alternatives. This efficiency is further validated in the end-to-end breakdown on Qwen2.5-7B (Figure~\ref{fig:gemm_bench}\textbf{(b)}), where ARCQuant results in a modest 4.9\% total latency increase. Notably, the specific cost of our Fused Quantization Kernel remains minimal, confirming that our design effectively maps the complexity of residual compensation into efficient, unified-precision GEMM operations.

\paragraph{End-to-End Inference Performance.} We evaluate prefill latency and memory on NVIDIA RTX PRO 6000 and RTX 5090. Detailed experimental results across varying sequence lengths are provided in Appendix~\ref{appendix:details}. Figure~\ref{fig:e2e} confirms substantial gains over FP16: Qwen2.5-7B achieves 2.0$\times$--2.5$\times$ speedup (PRO 6000) and Llama 3.1-8B reaches 3.5$\times$ (RTX 5090), while memory usage drops by 1.5$\times$--2.8$\times$. Crucially, residual overhead is marginal: compared to uncompensated NVFP4, latency increases by only 3\%--9\%.

\begin{table}[t]
\centering
\caption{End-to-end generation throughput in vLLM on Qwen2.5-7B using an RTX 5090 (batch size 8, generation length 128).}
\label{tab:vllm_gen}
\setlength{\tabcolsep}{3.5pt}
\begin{tabular}{l|cc|cc}
\toprule
& \multicolumn{2}{c|}{\textbf{SeqLen 1024}} & \multicolumn{2}{c}{\textbf{SeqLen 2048}} \\
\cmidrule(lr){2-3} \cmidrule(lr){4-5}
\textbf{Method} & Total & Decode & Total & Decode \\
\midrule
FP16 & 10756 & 1195 & 10203 & 600 \\
FP8 & 16699 & 1855 & 16857 & 992 \\
\rowcolor{lightpink}
\textbf{ARCQuant} & \textbf{21077} & \textbf{2342} & \textbf{21237} & \textbf{1249} \\
\bottomrule
\end{tabular}
\end{table}

\paragraph{Generation Throughput in vLLM.}
Since decoding throughput is a critical bottleneck in practical LLM serving, we further integrate ARCQuant into the vLLM engine and evaluate end-to-end generation performance on Qwen2.5-7B. As shown in Table~\ref{tab:vllm_gen}, we measure both total throughput and decoding throughput on an RTX 5090 with batch size 8 and generation length 128. ARCQuant consistently achieves the best performance across different context lengths. At sequence length 1024, ARCQuant reaches 21077.0 total tokens/s and 2341.9 decode tokens/s, corresponding to a 1.96$\times$ decoding speedup over FP16. At sequence length 2048, it achieves 21236.6 total tokens/s and 1249.2 decode tokens/s, yielding a 2.08$\times$ decoding speedup over FP16. In both settings, ARCQuant also substantially outperforms the FP8 reference, demonstrating that its efficiency advantages extend beyond prefill and remain effective in decoding generation scenarios.

\subsection{Ablation Studies}


\paragraph{Generalizability.} Although tailored for the fine-grained NVFP4 format, ARCQuant demonstrates robust extensibility to INT4 and MXFP4 formats. On Llama 3.1-8B, our method consistently outperforms RTN baselines across all metrics as detailed in Appendix~\ref{appendix:details}. Notably, ARCQuant reduces INT4 perplexity by 0.89 and improves zero-shot accuracy by 2.12\%. These gains indicate that while the dual-stage mechanism achieves optimal synergy with NVFP4, the underlying residual compensation principle remains effective in enhancing representation fidelity across various low-bit formats.

\paragraph{Calibration Robustness.} We evaluate the sensitivity of ARCQuant to calibration data distributions in two scenarios. First, using standard text-domain calibration (WikiText2) for Qwen2.5-Coder-7B-Instruct, we observe robust transfer to coding tasks (HumanEval, MBPP, and their extended versions), where ARCQuant retains over 99\% of FP16 accuracy and significantly outperforms Atom (Table~\ref{tab:code_results}). Second, varying the calibration source (WikiText2, C4, HumanEval)~\citep{wikitext, c4, humaneval} on Llama 3.1-8B results in marginal performance fluctuations ($<0.03$ in PPL and $<0.03\%$ in zero-shot accuracy) (see Appendix~\ref{appendix:details}). These results indicate that the outlier structures targeted by ARCQuant are stable, rendering the method robust to calibration data selection.

\paragraph{Scalability Across Model Sizes and Architectures.} 
We further evaluate ARCQuant on larger-scale and structurally different models to verify its applicability beyond standard dense 7B/8B settings. On Llama 3.1-70B, ARCQuant consistently improves over the NVFP4+RTN baseline across all reported metrics, with gains of 0.65 in average zero-shot accuracy, 0.23 lower WikiText2 perplexity, and 0.50 higher MMLU, while maintaining a competitive margin to the FP16 baseline. We also validate ARCQuant on the MoE model Mixtral 8x7B-Instruct, where it again achieves consistent improvements over NVFP4+RTN, remaining close to FP16 performance. Detailed results are provided in Appendix~\ref{appendix:details}. These results indicate that ARCQuant remains effective not only when scaling to substantially larger dense models, but also when applied to sparse MoE architectures.

\label{sec:bibtex}



\section{Conclusion}

\label{sec:conclusion}

In this paper, we presented ARCQuant, a framework designed to bridge the gap between fine-grained NVFP4 quantization and the outlier-heavy nature of LLM activations. Unlike rotation-based or mixed-precision approaches that compromise block isolation or hardware uniformity, ARCQuant resolves the outlier challenge through an augmented residual channel strategy. Our theoretical analysis confirms that this dual-stage mechanism achieves a worst-case error bound comparable to standard MXFP8. We translated this theoretical advantage into practical speedups via custom fused kernels, which enable dynamic residual compensation with minimal overhead. Crucially, ARCQuant secures these accuracy gains while maintaining strict compatibility with unified-precision hardware. As architectures evolve toward lower precisions, our principle of "trading minimal compute dimensions for higher fidelity" offers a scalable, hardware-friendly pathway for efficient LLM inference. Future work will extend this methodology to sub-4-bit formats and larger-scale architectures.

\section*{Limitations}
\label{sec:limitations}

While ARCQuant offers a hardware-efficient pathway for NVFP4 inference, we identify the following limitations to be addressed in future work:

\paragraph{Integration with Advanced Weight Quantization.}
Our current framework focuses primarily on resolving the activation outlier challenge, utilizing a standard Round-to-Nearest (RTN) strategy for weights. While effective, ARCQuant is theoretically compatible with advanced compensation-based weight quantization algorithms (e.g., GPTQ or AWQ). Integrating these methods could further enhance model fidelity, particularly for scenarios requiring sub-4-bit weight compression.

\paragraph{Hardware Dependency.}
ARCQuant is explicitly co-designed with the NVIDIA Blackwell architecture to leverage its unified-precision Tensor Core instructions. While our algorithmic contributions are general, the practical throughput benefits are contingent on hardware native support for the NVFP4 format. On legacy architectures lacking block-scaled acceleration, the method serves primarily as a simulation of future performance rather than an immediate deployment solution.

\paragraph{Dependence on Offline Calibration.}
The channel reordering indices and the outlier count $S$ in ARCQuant are determined offline based on calibration data. Although our experiments demonstrate strong robustness across different calibration sets, this static approach inherently assumes that the distribution of outlier channels remains relatively consistent during inference. While this trade-off is necessary to avoid the high latency of runtime permutation search, it may theoretically limit adaptability to extreme out-of-distribution inputs compared to fully dynamic (albeit slower) strategies.


\section*{Acknowledgments}

This work is supported in part by the National Natural Science Foundation of China (Grant No. 62550068 and No. 62276188), and the Emerging Frontiers Cultivation Program of Tianjin University Interdisciplinary Center.


\bibliography{custom}

\clearpage

\appendix

\section{Block-scaled Numerical Formats}
\label{appendix:block_formats}

Recent hardware advances have introduced block-scaled data formats to bridge the gap between high compression rates and representational fidelity. Unlike traditional per-tensor quantization, these formats partition tensors into small, fixed-size blocks, where elements within each block share a common scaling factor. We summarize the key specifications of these formats in Table~\ref{tab:formats}.

\paragraph{OCP Microscaling Formats (MXFP).} 
The Open Compute Project (OCP) defines a family of Microscaling formats, including MXFP8, MXFP6, and MXFP4~\citep{rouhani2023microscalingdataformatsdeep}. 
As shown in Table~\ref{tab:formats}, these formats uniformly adopt a block size of $g=32$. 
Crucially, they utilize an 8-bit scale factor in E8M0 format (exponent only), which acts as a shared exponent for the block. 
This design allows the elements (ranging from 4-bit to 8-bit) to share a dynamic range while maintaining efficient hardware implementation.

\paragraph{NVIDIA NVFP4.} 
Distinct from the OCP standard, the NVFP4 format targets finer-grained quantization with a block size of $g=16$~\citep{nvidia2024nvfp4}. 
This reduced block size offers tighter isolation of outliers. 
However, unlike the exponent-only scales of MX formats, NVFP4 employs an E4M3 scale factor. 
Due to the limited dynamic range of E4M3 compared to E8M0, NVFP4 necessitates an additional, global FP32 tensor scale to align the block-wise values correctly. 
This hierarchical scaling structure (Element $\to$ Block Scale $\to$ Tensor Scale) is a unique characteristic of NVFP4 that requires specialized handling in software kernels.

\section{Experiment Details}
\label{appendix:details}

\subsection{Experiment Setup}

\begin{table}[h]
\centering
\caption{Quantization overhead and efficiency. We report the calibration latency, quantization time and model memory on RTX PRO 6000 GPU.}
\label{tab:quant_efficiency}
\begin{tabular}{l|ccc}
\toprule
\textbf{Models} & \textbf{Calib.} & \textbf{Quant.} & \textbf{Mem.} \\
 & (s) $\downarrow$ & (s) $\downarrow$ & (GB) $\downarrow$ \\
\midrule
Llama 3.1-8B  & 79.84& 9.15  & 4.75  \\
Qwen2.5-7B    & 89.66& 9.38  & 4.24  \\
Qwen2.5-32B   & 176.44& 43.89 & 19.57 \\
\bottomrule
\end{tabular}
\end{table}

We used 128 samples from the WikiText2~\citep{wikitext} dataset, each with a sequence length of 2048, as the calibration set. 
Based on the derived reordering indices and the layer-specific outlier counts $S$, we perform weight quantization. 
Table~\ref{tab:quant_efficiency} reports the calibration latency, the resulting quantized model size, and the total quantization time.

\subsection{Results Breakdown}

\paragraph{Calibration Robustness.}
To assess the robustness of ARCQuant to calibration data variations, we conducted an ablation study using three distinct datasets: C4~\citep{c4}, WikiText2~\citep{wikitext}, and HumanEval~\citep{humaneval}. 
To ensure a fair comparison, all calibration sets were controlled to have an equivalent sample size. 
Table~\ref{tab:calibration_robustness} details the resulting zero-shot accuracy and WikiText2 perplexity.

\begin{table*}[t]
\centering
\caption{Zero-shot accuracy and WikiText2 perplexity, with different calibration datasets(C4, HumanEval, WikiText2). }
\label{tab:calibration_robustness}
\begin{tabular}{l|cccccc|c}
\toprule
 & \multicolumn{6}{c|}{\textbf{Zero-shot Accuracy} ($\uparrow$)} & \textbf{PPL} ($\downarrow$) \\
\textbf{Calibration Set} & Arc-C & Hella & Lamba & PIQA & Wino & \textbf{Average} & WikiText2 \\
\midrule
\multicolumn{8}{c}{\textbf{Llama 3.1-8B}} \\
\midrule
C4         & 52.22 & 76.80 & 73.30 & 79.71 & 72.38 & 70.88 & 6.90 \\
HumanEval  & 52.65 & 77.57 & 72.75 & 79.60 & 71.82 & 70.88 & 6.89 \\
WikiText2       & 52.22 & 77.22 & 72.79 & 79.16 & 73.09 & 70.90 & 6.87 \\
\bottomrule
\end{tabular}
\end{table*}

\paragraph{INT4 and MXFP4 Evaluation.} 
To provide a comprehensive assessment of ARCQuant's generalizability, we report the full zero-shot accuracy breakdown and perplexity scores on Llama 3.1-8B in Table~\ref{tab:full_llama_results}. 
Detailed comparisons across five standard benchmarks (ARC-Challenge, HellaSwag, Lambada, PIQA, Winogrande) confirm that ARCQuant consistently outperforms the RTN baseline in both INT4 and MXFP4 formats.

\begin{table*}[t]
\centering
\caption{Detailed zero-shot accuracy and perplexity on Llama 3.1-8B under INT4 and MXFP4 formats. \textbf{Bold} indicates the best performance within each quantization group.}
\label{tab:full_llama_results}
\begin{tabular}{l|ccccc|c|c}
\toprule
 & \multicolumn{5}{c|}{\textbf{Zero-shot Accuracy} ($\uparrow$)} & \textbf{Avg.} & \textbf{PPL} ($\downarrow$) \\
\textbf{Method} & Arc-C & Hella & Lamba & PIQA & Wino & ($\uparrow$) & WikiText2 \\
\midrule
FP16  & 53.58 & 78.90 & 75.47 & 81.18 & 74.03 & 72.63 & 6.24 \\
\midrule \midrule
\multicolumn{8}{c}{INT4 Quantization} \\
\midrule
RTN           & 45.99 & 73.82 & 66.14 & 77.37 & 65.59 & 65.78 & 8.84 \\
\rowcolor{lightpink} \textbf{ARCQuant} & 47.10 & 74.73 & 70.33 & 78.51 & 68.82 & \textbf{67.90} & \textbf{7.95} \\
\midrule \midrule
\multicolumn{8}{c}{MXFP4 Quantization} \\
\midrule
RTN           & 47.44 & 75.01 & 69.40 & 78.67 & 69.61 & 68.03 & 7.86 \\
\rowcolor{lightpink} \textbf{ARCQuant} & 48.46 & 75.51 & 72.13 & 78.18 & 70.56 & \textbf{68.97} & \textbf{7.50} \\
\bottomrule
\end{tabular}
\end{table*}

\begin{figure}[t]
  \centering
  \includegraphics[width=\columnwidth]{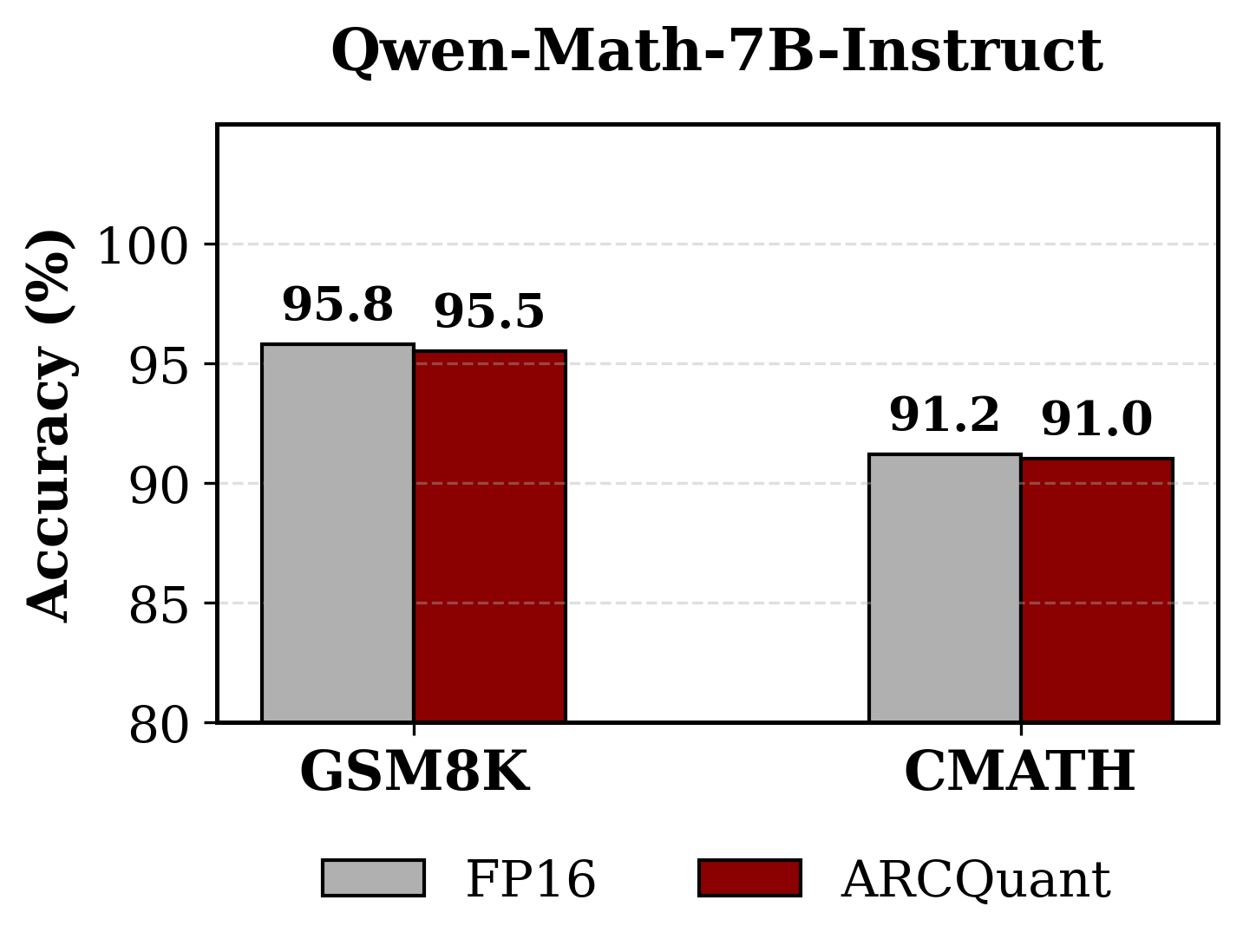}
  \caption{Qwen2.5-Math-7B-Instruct on GSM8K and CMATH}
  \label{fig:math_eval}
\end{figure}

\paragraph{Math Evaluation.} 
We evaluate mathematical reasoning capabilities using Qwen2.5-Math-7B-Instruct on the GSM8K and CMATH benchmarks. 
As shown in Figure~\ref{fig:math_eval}, ARCQuant retains over 99\% of the FP16 baseline accuracy, demonstrating exceptional robustness in domain-specific tasks.

\begin{table*}[t]
\centering
\caption{Comparison of parameters for MX-compliant formats and NVFP4. $d$ and $w$ denote the bit-width of elements and block scales, respectively. Notably, NVFP4 introduces an additional per-tensor scale in FP32.}
\label{tab:formats}
\begin{tabular}{l|c|l|c|c|c|c|c|c}
\toprule
\textbf{Format} & \textbf{Element} & \textbf{Element Data} & \textbf{Bias} & \textbf{Max} & \textbf{Block} & \textbf{Scale} & \textbf{Scale} & \textbf{Tensor} \\
\textbf{Name} & \textbf{Bits ($d$)} & \textbf{Type} & \textbf{($b$)} & \textbf{Normal} & \textbf{Size ($g$)} & \textbf{Type} & \textbf{Bits ($w$)} & \textbf{Scale} \\
\midrule
\multirow{2}{*}{MXFP8} & \multirow{2}{*}{8} & FP8 (E5M2) & 15 & $\pm$ 57344 & \multirow{2}{*}{32} & \multirow{2}{*}{E8M0} & \multirow{2}{*}{8} & \multirow{2}{*}{N/A} \\
                       &                    & FP8 (E4M3) & 7  & $\pm$ 448   &                     &                       &                    &                      \\
\midrule
\multirow{2}{*}{MXFP6} & \multirow{2}{*}{6} & FP6 (E3M2) & 3  & $\pm$ 28    & \multirow{2}{*}{32} & \multirow{2}{*}{E8M0} & \multirow{2}{*}{8} & \multirow{2}{*}{N/A} \\
                       &                    & FP6 (E2M3) & 1  & $\pm$ 7.5   &                     &                       &                    &                      \\
\midrule
MXFP4                  & 4                  & FP4 (E2M1) & 1  & $\pm$ 6     & 32                  & E8M0                  & 8                  & N/A                  \\
\midrule
NVFP4         & 4                  & FP4 (E2M1) & 1  & $\pm$ 6     & 16                  & E4M3                  & 8                  & FP32        \\
\bottomrule
\end{tabular}
\end{table*}

\paragraph{End-to-End Inference.}
To supplement the efficiency analysis in Section~\ref{sec:experiments}, we provide the detailed breakdown of prefill performance. 
Table~\ref{tab:prefill_performance_appendix} reports the latency and peak memory consumption across varying batch sizes and sequence lengths (512, 1024, 2048) on NVIDIA RTX PRO 6000 and RTX 5090 GPUs.

\paragraph{Larger-Scale and MoE Model Evaluation.} 
To further assess the scalability of ARCQuant across model sizes and architectures, we report the full zero-shot accuracy breakdown, perplexity, and MMLU results on Llama 3.1-70B and Mixtral 8x7B-Instruct in Table~\ref{tab:large_model_results}. 
Detailed comparisons confirm that ARCQuant consistently outperforms the NVFP4+RTN baseline on both the larger dense model and the sparse MoE model, demonstrating its effectiveness beyond standard 7B/8B dense LLM settings.

\begin{table*}[t]
\centering
\caption{Zero-shot, few-shot accuracy and perplexity. \textbf{Bold} indicates the best performance among NVFP4 methods.}
\label{tab:large_model_results}
\begin{tabular}{l|ccccc|c|c}
\toprule
 & \multicolumn{5}{c|}{\textbf{Zero-shot Accuracy} ($\uparrow$)} & \textbf{PPL} ($\downarrow$) & \textbf{MMLU} ($\uparrow$) \\
\textbf{Method} & Arc-C & Hella & Lamba & PIQA & \textbf{Average} & WikiText2 & 5-shot \\
\midrule
\multicolumn{8}{c}{\textbf{Llama 3.1-70B}} \\
\midrule
FP16       & 64.85 & 84.99 & 78.79 & 84.17 & 78.20 & 2.81 & 78.57 \\
NVFP4 + RTN         & 59.30 & 83.90 & 75.88 & 82.64 & 75.43 & 3.85 & 76.11 \\
\rowcolor{lightpink} \textbf{ARCQuant} & \textbf{59.56} & \textbf{84.28} & \textbf{76.98} & \textbf{83.51} & \textbf{76.08} & \textbf{3.62} & \textbf{76.61} \\

\midrule \midrule 

\multicolumn{8}{c}{\textbf{Mixtral 8x7B-Instruct}} \\
\midrule
FP16       & 65.61 & 86.00 & 77.29 & 84.93 & 78.46 & 4.14 & 70.30 \\
NVFP4 + RTN         & 64.16 & \textbf{85.09} & 77.55 & 83.73 & 77.63 & 4.41 & 68.54 \\
\rowcolor{lightpink} \textbf{ARCQuant} & \textbf{65.10} & 84.94 & \textbf{78.03} & \textbf{84.06} & \textbf{78.03} & \textbf{4.38} & \textbf{68.71} \\
\bottomrule
\end{tabular}
\end{table*}

\section{Reproducibility Statement}
\label{appendix:reproducibility}

We are committed to facilitating the reproduction of our results. 

\paragraph{Code and Data.}
Our source code, compatible with PyTorch 2.9.0 and CUDA 12.8, is available at the GitHub repository: \url{https://github.com/actypedef/ARCQuant}. 
To facilitate reproduction, we provide the calibration artifacts in the supplementary material. 
We utilize publicly available datasets and official model checkpoints from Hugging Face without modification.
\paragraph{Implementation Details.}
For calibration, we randomly select 128 segments from WikiText2 with a sequence length of 2048. The outlier threshold is set to $\tau = 2^{-3}M$. To guarantee deterministic behavior, we fix the random seed to 0 across all experiments. 
\paragraph{Baselines and Hardware:}
Baselines (e.g., Atom, FlatQuant) are reproduced using their official repositories. All efficiency benchmarks reported in Section~\ref{sec:experiments} were conducted on NVIDIA RTX 5090 and RTX PRO 6000 GPUs.

\section{Kernel Implementation Details}
\label{appendix:kernel_design}

\paragraph{Interleaved Channel Layout.}
While our mathematical formulation in Section~\ref{sec:methodology} denotes the augmented matrix as a simple concatenation ($[Q_X \mid Q_{R_{\text{o}}}]$), implementing this logically contiguous layout directly would incur significant latency due to strided global memory access patterns.
To overcome this, we designed a hardware-friendly Interleaved Channel Layout for the actual kernel implementation.
Specifically, for the $S$ outlier channels requiring compensation, we group them into blocks of 16 (matching the NVFP4 block size).
Instead of separating all main quantizations and residual quantizations into distant memory regions, we interleave them locally: a 16-channel primary block is immediately followed by its corresponding 16-channel residual block in physical memory.
This layout allows our Fused Quantization Kernel to compute both the primary and residual quantization values within on-chip registers and perform continuous, coalesced write-back operations to global memory.
The weight matrix $W$ is pre-processed offline to strictly match this interleaved pattern, ensuring that the standard GEMM operation yields the correct mathematical result without modification.

\paragraph{Compatibility and Future Work.}
As discussed in the Limitations, ARCQuant represents a quantization strategy for fine-grained formats rather than a hardware-specific workaround.
As block-scaled formats like NVFP4 and MXFP4 gain wider adoption, the utility of our residual compensation mechanism becomes increasingly significant.
A key advantage of our design is the decoupling of quantization logic from computation.
Since the error compensation is mapped entirely into the input data space (via channel augmentation), ARCQuant is compatible with any standard, high-performance GEMM kernel (e.g., cuBLAS or CUTLASS) without requiring modifications to the inner loop of the matrix multiplication.
Adapting ARCQuant to other emerging formats simply requires updating the Fused Quantization Kernel to support the target format's encoding, preserving the efficiency of the downstream GEMM.
Finally, while ARCQuant has already been integrated into vLLM for end-to-end benchmarking, we plan to further develop full serving support for high-throughput engines such as vLLM and validate its broader deployment potential.

\begin{table*}[t]
\centering
\caption{Detailed prefill performance across various models and hardware configurations. We report the prefill latency (ms) and peak memory consumption (GB) for ARCQuant, FP16, and NVFP4. Bsz denotes batch size and Len denotes sequence length.}
\label{tab:prefill_performance_appendix}
\begin{tabular}{ll|cc|cc|cc}
\toprule
\textbf{Setting} & \textbf{Model} & \multicolumn{2}{c|}{\textbf{ARCQuant}} & \multicolumn{2}{c|}{\textbf{FP16}} & \multicolumn{2}{c}{\textbf{NVFP4}} \\
(Bsz / Len) & & Latency& Memory& Latency& Memory& Latency& Memory\\
 & & (ms)& (GB)& (ms)& (GB)& (ms)&(GB)\\
\midrule
\multicolumn{8}{c}{Hardware: NVIDIA RTX PRO 6000} \\
\midrule
32 / 512  & Qwen2.5-7B  & 440.49 & 12.76 & 881.37  & 29.16 & 431.21  & 12.53 \\
32 / 1024 & Qwen2.5-7B  & 881.72 & 20.87 & 1895.40 & 44.08 & 857.81  & 20.64 \\
32 / 2048 & Qwen2.5-7B  & 1756.42& 37.10 & 4162.91 & 73.92 & 1707.77 & 36.86 \\
\midrule
16 / 512  & Qwen2.5-14B & 393.66 & 17.73 & 885.61  & 36.22 & 383.10  & 17.28 \\
16 / 1024 & Qwen2.5-14B & 793.21 & 26.05 & 1830.07 & 44.79 & 767.21  & 25.60 \\
16 / 2048 & Qwen2.5-14B & 1592.07& 42.67 & 4048.28 & 61.91 & 1544.15 & 42.22 \\
\midrule
8 / 512   & Qwen2.5-32B & 445.24 & 26.39 & 902.02  & 65.57 & 421.01  & 24.60 \\
8 / 1024  & Qwen2.5-32B & 877.41 & 32.17 & 1844.20 & 70.11 & 830.58  & 30.38 \\
8 / 2048  & Qwen2.5-32B & 1761.93& 43.74 & 3908.75 & 79.18 & 1668.84 & 41.95 \\

\midrule \midrule

\multicolumn{8}{c}{Hardware: NVIDIA RTX 5090} \\
\midrule
4 / 512   & Llama 3.1-8B  & 72.72  & 6.67  & 253.05  & 16.70 & 67.51   & 6.08  \\
4 / 1024  & Llama 3.1-8B  & 134.57 & 7.87  & 468.70  & 18.44 & 123.60  & 7.28  \\
4 / 2048  & Llama 3.1-8B  & 270.50 & 10.27 & 917.05  & 21.92 & 246.68  & 9.68  \\
\midrule
4 / 512   & Qwen2.5-7B  & 69.27  & 5.65  & 226.83  & 16.10 & 66.95   & 5.44  \\
4 / 1024  & Qwen2.5-7B  & 128.17 & 6.67  & 414.51  & 17.97 & 122.20  & 6.45  \\
4 / 2048  & Qwen2.5-7B  & 251.33 & 8.70  & 888.14  & 21.70 & 237.60  & 8.48  \\
\bottomrule
\end{tabular}
\end{table*}

\end{document}